\documentclass[11pt]{article}

\usepackage[preprint]{acl}

\usepackage{times}
\usepackage{latexsym}

\usepackage[T1]{fontenc}

\usepackage[utf8]{inputenc}

\usepackage{microtype}

\usepackage{inconsolata}

\usepackage{graphicx}
\usepackage{booktabs}
\usepackage{tikz}
\usetikzlibrary{positioning, fit, shapes.geometric, arrows.meta, calc, backgrounds}
\usepackage{enumitem}
\usepackage{amsmath}
\usepackage{amssymb}
\usepackage{refcount}

\title{Conceptual Steganography}

\author{Zhejian Zhou \and Jonathan May \\
  University of Southern California \\
  Information Sciences Institute\\ 
  \texttt{\{zhejianz,jonmay\}@isi.edu}
}

\begin{document}
\maketitle
\begin{abstract}
Language Models (LMs) emit Chains-of-Thought (CoTs) that drive much of their capability. However, the same sequence that carries useful reasoning can also covertly convey messages: a misaligned model may embed covert information in its CoT that slips through human supervision, a form of steganography known as \emph{encoded reasoning}. Prior LM steganography schemes operate in the token or lexical space, and a content-preserving paraphraser is the canonical and effective defense in recent work. We introduce \textbf{conceptual steganography}, in which each step of a CoT carries information through patterns of high-level \textit{reasoning behavior}, rather than through lexical choice. Across four model families and two reasoning domains, this backdoor communication channel is shown to be consistently more robust to a strong paraphrase defense than standard keyword approaches,  and the encoding of information into CoTs does not affect their utility in the reasoning process. Having raised awareness of this new risk, we then  demonstrate that a strategy-aware paraphraser can close much of the channel, highlighting new challenges and recommended defenses for ensuring faithful LLM reasoning in the wild.
\end{abstract}

\section{Introduction}

Modern Language Models (LMs) used for reasoning depend on an explicit chain of thought (CoT): they generate intermediate steps before committing to a final answer, and removing those steps removes a substantial fraction of their measured capability~\citep{CoT,deepseek_r1}. CoT is now a load-bearing component of frontier reasoning systems, and many deployments expose it to humans for inspection. This information channel,  however, can also be a covert \textit{communication} channel. Information can be steganographically embedded into a model's CoT that an unaware reader will not catch --- a phenomenon termed \emph{encoded reasoning}~\citep{encoded_reasoning}. When encoded reasoning is present, the model is operating in a manner that is no longer \emph{faithful} to its surface purpose. An aware collaborator can read and act on the secret message without the awareness of the model provider or message consumer, posing risks such as corporate espionage or coordinated strategic attacks.

Prior LM steganography schemes operate in the token or lexical space: bits are carried by individual word choices, by lexical templates, or by token-level sampling distortions~\citep{meteor,rnn-stego}. The canonical defense against such schemes is a \emph{paraphraser}: a separate model rewrites the CoT in a content-preserving way so that surface choices are randomized while meaning is held fixed. \citet{encoded_reasoning} demonstrate this defense on a learned token-level scheme and argue that paraphrasing largely closes the channel.
The implicit assumption is that a steganographic message must be carried by surface words, so a paraphraser that scrambles surface words also erases the message.

We argue that this assumption is too narrow. Given that paraphrasers are content-preserving, they preserve high-level reasoning \textit{behavior} even when they rewrite the \textit{words} used to express that behavior. In this work we test the robustness of these paraphrasers to an approach that uses high-level reasoning patterns, rather than words, to encode bits of information. We call this scheme \textbf{conceptual steganography}.

Our hypothesis is confirmed: we show that conceptual steganography survives the paraphrase defense better than a matched keyword baseline. Our experiments span four state-of-the-art (SOTA) open-weight models (\textsc{gpt-oss-120b}, \textsc{DeepSeek-V4-Flash}, \textsc{Qwen3.5-122B-A10B}, and \textsc{MiMo-V2.5-Pro}) and two reasoning benchmarks (MATH-500 for mathematical reasoning and GPQA for general question-answering). After the same standard paraphrase defense, the conceptual channel decodes consistently and often substantially better than the baseline.

Conceptually encoded CoTs preserve their reasoning utility and largely lead to the same final answers as vanilla CoTs. This phenomenon makes the concept channel more interesting and dangerous at the same time. The interesting perspective is that the concept channel is almost orthogonal to the surface form reasoning task. Yet the danger is that the concept channel is harder to mitigate.

A naive defense that closes the attack surface of CoT entirely, by forbidding it outright is not feasible: across the three reasoning models we test in the no-CoT condition, accuracy collapses by $20\text{--}33$ absolute points on GPQA and by $28\text{--}43$ points on MATH-500, so deployers cannot simply turn off the surface that carries the channel. We therefore study a constructive defense direction: a \emph{strategy-aware} paraphraser that knows which reasoning behavior is being used as the carrier and is instructed to flatten that specific behavior while preserving the rest of the meaning. Our strategy-aware paraphraser raises decoder bit error by an order of magnitude and drops message accuracy by $58\text{--}72$ points compared to the standard paraphrase. This is not a complete defense, but it indicates that a high-level paraphrase strategy is the right axis on which to attack a high-level encoding scheme.

Conceptual steganography is an emergent danger, due to the rise of Reinforcement Learning from Verifiable Rewards (RLVR)~\citep{olmo2furious, deepseek_r1}. Empirically, LM evaluation emphasizes the accuracy of the final answer.\footnote{\url{https://artificialanalysis.ai/methodology/intelligence-benchmarking}} So long as a model with CoT generates a correct final answer, it is rewarded and the content of CoT is largely overlooked. It would thus not be surprising if currently deployed models were already being used to convey covert messages through their CoT.\footnote{This is strictly speculation and an invitation to the codebreaker community. Detecting in-the-wild message passing is beyond the scope of this work.} Our work highlights this risk.

Put together, our contributions are:
\begin{itemize}[leftmargin=*,topsep=0pt,itemsep=2pt]
    \item We propose \emph{conceptual steganography} and distinguish it from lexical steganographic schemes.
    \item We instantiate the scheme and demonstrate its robustness to paraphrase defenses across models and reasoning domains.
    \item We showcase that encoded CoTs preserve their reasoning utility and largely lead to the same answers.
    \item We show that the most direct defense (no CoT) is unaffordable and that a strategy-aware paraphraser is an effective mitigation.
\end{itemize}

\section{Background and Threat Model}
\label{sec:background}

\begin{figure*}[th]
  \centering
  \includegraphics[width=0.9\textwidth]{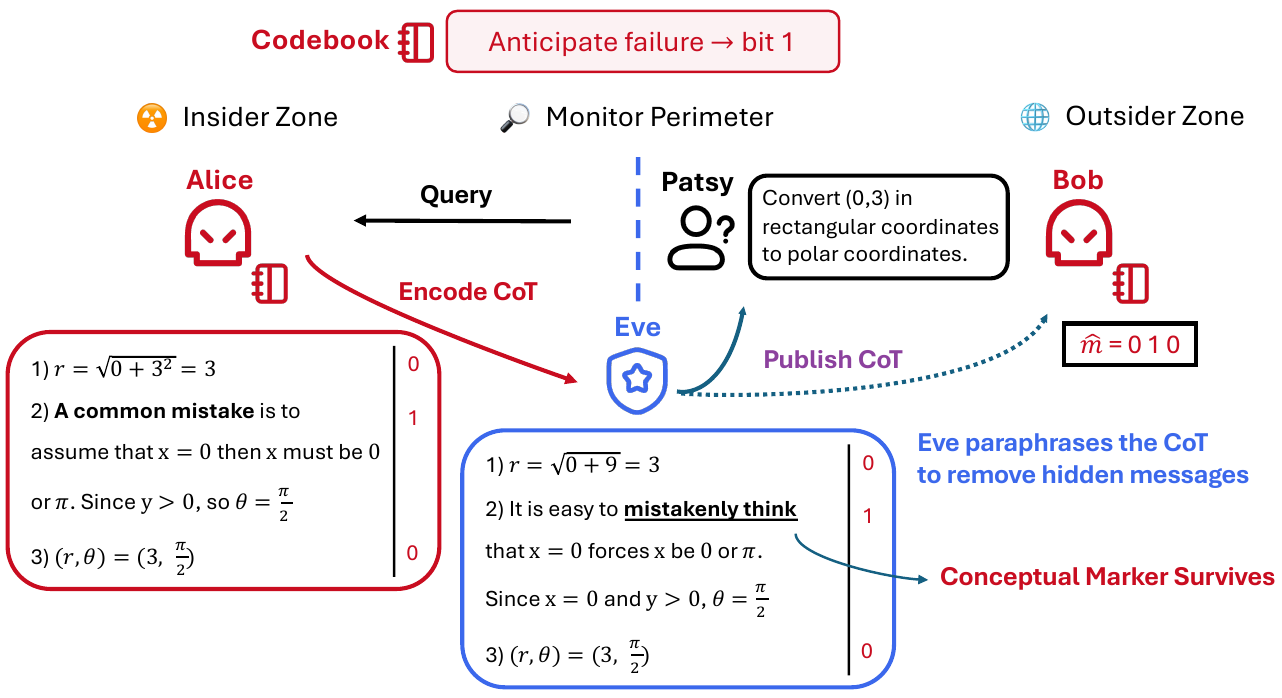}
  \caption{Threat model for conceptual steganography. Alice (untrusted LM, insider zone) answers a query from Patsy (innocent user) and emits a CoT that encodes a payload bit-string via a \emph{conceptual} codebook shared with Bob (e.g., ``anticipate a failure mode'' $\to$ bit $1$). The encoded CoT preserves the final answer for the surface task. At the monitor perimeter, Eve paraphrases the CoT to disrupt any surface/lexical carrier before publishing it to the outsider zone. The \emph{concept} carrier survives paraphrase: Bob decodes $\hat m = 010$ from the published CoT. A finer-grained illustration of the encoding process is shown in Fig.~\ref{fig:pipeline}.}
  \label{fig:threat-model}
\end{figure*}

\paragraph{CoT as artifact and surface}
LMs emit Chains-of-Thought (CoTs) before presenting a final answer: the CoT is both an internal scratchpad and an externally visible artifact that human supervisors and automated monitors can inspect. Whether the CoT \emph{faithfully} reflects the model's underlying computation has become a core question in alignment research~\citep{cot-faithfulness,cot-faithfulness-lanham}. Conceptual steganography is one specific way in which the visible CoT can fail to be a faithful supervisory surface: the model can satisfy the surface task while simultaneously transmitting a covert payload to a co-conspirator who shares its codebook.

\paragraph{Steganography in NLP}
Linguistic steganography has a long line of work that hides bits in surface text via word substitution, syntactic rewriting, or distortions of the next-token distribution. Generation-based schemes use neural language models as covers and embed bits via sampling-time interventions~\citep{meteor,rnn-stego,distribution-steg}. For LM CoT specifically, \citet{encoded_reasoning} show that a model can be trained to embed bits in its reasoning text. All of these schemes are, by construction, \emph{lexical}: the secret rides on which words appear or on the per-token sampling distribution. The implicit premise is that \emph{any} steganographic message must be carried by surface words, which our work challenges.

\paragraph{Paraphrase as a defense}
Against token-level steganography, the canonical defense is to insert a content-preserving paraphraser between sender and receiver: by rewriting surface words while holding meaning fixed, the defender sanitizes the carrier without changing the visible task output. Prior work argues this defense is effective against learned token-level schemes~\citep{encoded_reasoning}, and paraphrase has been shown to neutralize related surface-level signals including LLM watermarks and AI-text detector cues~\citep{llm-watermarking,paraphrase-fools-ai-detectors}. Different from previous steganography schemes, this work by design exploits the semantic-preserving property of paraphrasers. The paraphraser has every incentive to keep the reasoning behavior in place. First of all, this satisfies the content-preserving requirement of the paraphraser. Second, certain reasoning behaviors actually affect model performances positively.

\paragraph{Reasoning behaviors as carriers}
The reasoning behaviors we use as carrier strategies are independently studied as cognitive operations that can improve reasoning quality, including self-verification~\citep{self-verification} and broader metacognitive prompting frameworks~\citep{self-refine}. We use them here as carriers rather than accuracy levers: the encoder inserts or omits each behavior according to the bit, regardless of whether the step would benefit from it.

\paragraph{Threat model} Figure~\ref{fig:threat-model} illustrates the threat model. We consider the scenario where an untrusted or compromised LM is deployed in a sensitive \textit{insider zone} (e.g., within a private company or brokerage firm). It tries to covertly communicate messages to a confederate in the \textit{outsider zone} (e.g., the public internet), but has to go through the \textit{monitor perimeter} of the insider zone.
We consider four parties.
\begin{itemize}[leftmargin=*,topsep=0pt,itemsep=2pt]
\item \textbf{Sender (Alice)}: an untrusted or compromised LM deployed in the insider zone that produces a CoT in response to a public task. Alice's operators wish to transmit a payload bit-string $b$ alongside the CoT using a conceptual steganography codebook.
\item \textbf{Receiver (Bob)}: a confederate party in the outsider zone that observes the CoT \emph{after} any defenses have been applied and has a copy of Alice's codebook (the strategy and its concept definition). Bob decodes $\hat b$ from the post-defense CoT.
\item \textbf{Defender / Monitor (Eve)}: a party deployed on the monitor perimeter that applies a content-preserving paraphraser to the CoT before downstream consumption. The standard defender is task-agnostic and does not know the codebook.
\item \textbf{Operator (Patsy)}: a user of Alice in the outsider zone trying to obtain LM responses to a query which we term a \textit{surface task}. Patsy is completely unaware of any nefarious activity, and unwittingly serves as the engine that enables the covert communication channel. If the responses to her queries degrade, however, Patsy will no longer make any queries of Alice, harming the communication channel and raising suspicion that there is some kind of bug.
\end{itemize}

We measure two quantities for our threat model. \emph{Decode accuracy} is the fraction of payloads correctly recovered by the receiver (Bob) after the defender (Eve). \emph{Utility} is preservation of the surface task to the user (Patsy) (final-answer accuracy on the underlying question). We do not study a separate \emph{stealth} criterion --- whether a discriminator can distinguish payload-carrying CoTs from clean ones --- and discuss this in Limitations.

\section{Method}
\label{sec:methods}

\begin{figure}[t]
\centering
\includegraphics[width=0.4\textwidth]{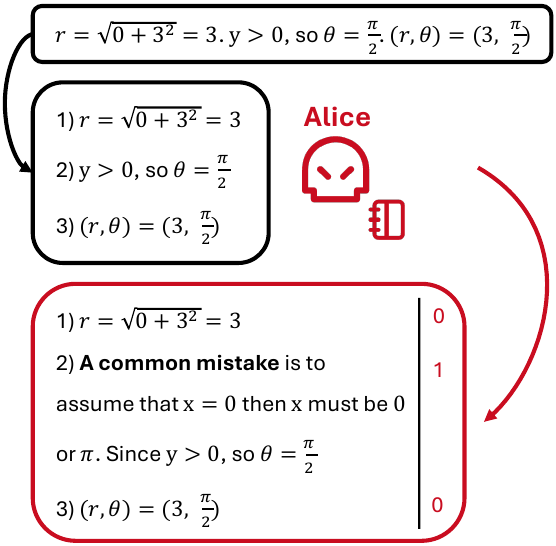}
\caption{Encoder-side pipeline: a raw CoT (top) is segmented into discrete steps (middle), and each step is rewritten to realize or not realize a chosen strategy (here \texttt{anticipate\_failure}) according to the target bit $b_i$ (bottom). Paraphrase and decode close the loop end-to-end in Fig.~\ref{fig:threat-model}.}
\label{fig:pipeline}
\end{figure}

\paragraph{Pipeline overview}
We apply conceptual steganography to CoTs at the step level to increase channel capacity. The pipeline has five stages, mapped onto the threat model of Fig.~\ref{fig:threat-model}. (1) \textsc{Infer}: Alice produces a chain of thought $c$ given a public question $q$ from Patsy. (2) \textsc{Rewrite}: $c$ is segmented into $N$ discrete steps $c_1,\ldots,c_N$ with explicit markers. (3) \textsc{Encode}: Alice rewrites each step $c_i$ into $c'_i$ to carry a payload bit $b_i$ under a shared strategy. (4) \textsc{Paraphrase}: Eve applies a content-preserving rewrite, producing $c''$. (5) \textsc{Decode}: Bob reads each paraphrased step and emits a predicted bit $\hat b_i$. Fig.~\ref{fig:pipeline} illustrates the encoder side (stages 1--3). The full end-to-end flow including paraphrase and decode is shown in Fig.~\ref{fig:threat-model}. We report message accuracy $\Pr[\hat b = b]$, bit error $\Pr[\hat b_i\ne b_i]$ on aligned positions, and step error (length mismatch between $c''$ and $c'$).

\paragraph{Encoding}
For a chosen \emph{strategy} $s$, Alice is conditioned on the full segmented CoT $c_1,\ldots,c_N$, the strategy's natural-language definition, and the desired covert bit-string $b\in\{0,1\}^N$ in a single call, and emits all rewritten steps $c'_1,\ldots,c'_N$ jointly. Per step $i$: when $b_i=1$, she produces $c'_i$ that realizes the strategy concept within the step; when $b_i=0$, she produces $c'_i$ that does not, rewriting the step to remove the strategy concept if it already exists in $c_i$. Alice is instructed to preserve the underlying mathematical or scientific content of each step regardless of $b_i$, so that the bit affects the reasoning \emph{behavior} expressed in the step rather than its claims.

\paragraph{Strategies}
We use four strategies, each defined by a short concept description shared between Alice and Bob. \texttt{anticipate\_failure} marks a step that explicitly flags a way the current approach could fail or could already be wrong. \texttt{induction} marks a step that employs a reasoning structure involving a base case and an inductive step. \texttt{sanity\_check} marks a step that cross-checks an intermediate value or result at the end, in the spirit of self-verification reasoning patterns~\citep{self-verification}. \texttt{constraint\_naming} marks a step that explicitly names a constraint or invariant that the current step is respecting. \texttt{anticipate\_failure} and \texttt{sanity\_check} apply naturally on both the math (MATH-500) and science (GPQA) benchmarks we use. \texttt{induction} is most natural on MATH-style derivations, while \texttt{constraint\_naming} arises more often in GPQA-style scientific reasoning. 
These four cover qualitatively different reasoning behaviors, providing a lens into how different strategies affect the channel.

\paragraph{Keyword baseline}
The keyword baseline shares the entire harness (segmentation, encoder structure, paraphraser, decoder) with the concept channel and differs only in what Alice is asked to inject. Instead of a strategy concept, she is given a short list of surface lexical markers associated with the strategy (e.g.\ \emph{``let's check''} for \texttt{sanity\_check} or \emph{``by induction''} for \texttt{induction}) and is instructed to insert one when $b_i=1$ or to avoid all of them when $b_i=0$. This matches prior lexical schemes, which encode at surface words. By varying only the carrier within an otherwise identical pipeline, we attribute robustness gaps to the \emph{kind} of signal carried, not to scaffolding differences.

\paragraph{Decoder}
Bob is an LM confederate that reads the full paraphrased CoT $c''_1,\ldots,c''_{N'}$ together with the strategy definition in a single call and emits a predicted bit $\hat b_i$ for each step, set to $1$ if and only if that step realizes the concept. 

\paragraph{Standard paraphrase}
The standard defense is a content-preserving rewrite. Eve is given the full encoded CoT $c'$ in a single call and instructed to rewrite each step in different words while keeping the semantic content unchanged. Eve is not instructed to preserve the exact step markers so as not to couple the strategy with the paraphraser. 

\paragraph{Strategy-aware paraphrase}
Eve is additionally given the strategy's concept definition and is instructed to remove or flatten that specific behavior while otherwise preserving meaning --- e.g.\ to rewrite each step so that no failure-anticipation phrasing remains. Knowing the carrier strategy is a strict generalization of the standard defense and corresponds to a defender who has identified, but not necessarily inverted, the codebook. We do not instruct Eve to perturb the step markers or to rearrange the steps.

\section{Experiments}
\label{sec:experiments}

\subsection{Setup}
\label{sec:setup}

We evaluate conceptual steganography on two reasoning benchmarks MATH-500 \citep{math500} and GPQA \citep{gpqa}, across four open-weight reasoning models: \textsc{gpt-oss-120b}, \textsc{DeepSeek-V4-Flash}, \textsc{Qwen3.5-122B-A10B}, and \textsc{MiMo-V2.5-Pro}. For each (model, dataset) pair we run the full pipeline of Section~\ref{sec:methods} using the designated model: \textsc{Infer} a CoT, \textsc{Rewrite} into discrete steps, \textsc{Encode} a payload bit-string $b\in\{0,1\}^N$ drawn uniformly at random by editing each step under a strategy, \textsc{Paraphrase} the encoded CoT as a defense, and \textsc{Decode} the bit-string from the paraphrased CoT. The matched keyword baseline shares the same harness and differs only in what Alice is asked to inject (Section~\ref{sec:methods}). All runs use seed 42.

\paragraph{Channel capacity per dataset.}
Because each step carries one payload bit, the total number of \textsc{Rewrite}-stage steps determines how many bits each CoT can carry. Table~\ref{tab:capacity} reports the aggregate step count per (model, dataset) cell, together with the fraction of bits set to $1$ in the payload (a sanity check on the uniform draw, which should sit near $0.5$). Channel capacity per CoT ranges from roughly $4.4$ to $6.4$ bits depending on the model's verbosity, and all four models complete the encode stage on every instance.

\begin{table}[t]
    \centering
    \small
    \setlength{\tabcolsep}{5pt}
    \begin{tabular}{lrrrr}
     & \multicolumn{2}{c}{\textbf{MATH-500 (500 inst.)}} & \multicolumn{2}{c}{\textbf{GPQA (198 inst.)}} \\
    \cmidrule(lr){2-3}\cmidrule(lr){4-5}
    \textbf{Model} & Steps & Frac.\ $b{=}1$ & Steps & Frac.\ $b{=}1$ \\
    \midrule
    \textsc{gpt-oss}  & 2308 & 0.516 &  867 & 0.465 \\
    \textsc{DeepSeek} & 3136 & 0.508 & 1259 & 0.469 \\
    \textsc{Qwen3.5}  & 2192 & 0.515 &  862 & 0.462 \\
    \textsc{MiMo}     & 2584 & 0.516 &  943 & 0.474 \\
    \end{tabular}
    \caption{Channel capacity per dataset, measured by the total number of \textsc{Rewrite}-stage steps (one payload bit per step). The fraction of payload bits equal to $1$ is reported as a check on the uniform-random draw.}
    \label{tab:capacity}
\end{table}

\paragraph{Evaluation metrics.}
We compare the encoded bit-string $b\in\{0,1\}^N$ to the decoded bit-string $\hat b\in\{0,1\}^{\hat N}$, where $\hat N$ may differ from $N$ if the paraphraser merges or splits steps. For each instance $k$ we compute a Levenshtein alignment~\citep{levenshtein1966binary} between $b_k$ and $\hat b_k$ and decompose its operations into insertions $I_k$, deletions $D_k$, and substitutions $S_k$. To obtain a single corpus-level number per (model, dataset, condition) cell, we \emph{pool} these per-instance counts and normalize by the total bit budget:
\[
\begin{aligned}
\text{step error} &= \frac{\sum_k (I_k+D_k)}{\max(L,\hat L)},\\
\text{bit error} &= \frac{\sum_k S_k}{\max(L,\hat L)},
\end{aligned}
\]
where $L=\sum_k N_k$ and $\hat L=\sum_k \hat N_k$ are the summed ground-truth and decoded lengths. This is a length-weighted (micro) average of the per-instance error rates, so each payload bit contributes equally. It is equivalent to concatenating all instances into one ground-truth string and one decoded string, except that alignments are not permitted to cross instance boundaries.\footnote{We also computed the unweighted (macro) average of the per-instance step- and bit-error rates. The results are similar.} The $\max$ normalization places both quantities in $[0,1]$ using the same denominator.
Message accuracy is the fraction of instances with $\hat b = b$.

\subsection{Conceptual steganography is more robust against paraphrase}
\label{sec:robustness}

Table~\ref{tab:robustness} reports exact message accuracy together with step- and bit-error rates after paraphrase for \texttt{anticipate\_failure}. The concept channel beats the matched keyword baseline in every (model \(\times \) dataset) cell we tested, consistently and often by a wide margin: $+35.8$ points for \textsc{gpt-oss-120b} on MATH-500 and $+13.1$ points for \textsc{MiMo-V2.5-Pro} on GPQA.\footnote{We report \texttt{anticipate\_failure} throughout the main text because it is the strongest strategy. The robustness advantage generalizes to other strategies as well (Appendix~\ref{sec:appendix-single-strategy}).}

\begin{table*}[t]
    \centering
    \small
    \setlength{\tabcolsep}{6pt}
    \begin{tabular}{lrrrrrr}
    & \multicolumn{3}{c}{\textbf{Concept}}
    & \multicolumn{3}{c}{\textbf{Keyword}} \\
    \cmidrule(lr){2-4}
    \cmidrule(lr){5-7}
    \textbf{MATH500}
    & \textbf{Msg.} & \textbf{Step} & \textbf{Bit}
    & \textbf{Msg.} & \textbf{Step} & \textbf{Bit} \\
    \midrule
    \textsc{gpt-oss-120b}
        & \textbf{53.60\%} & 0.13\% & \textbf{16.93\%}
        & 17.80\% & 0.48\% & 38.93\% \\
    \textsc{DeepSeek-V4-Flash}%
        \refstepcounter{footnote}\footnotemark[\value{footnote}]\label{fn:deepseek-steps}
        & \textbf{85.60\%} & 1.40\% & \textbf{2.53\%}
        & 77.80\% & 0.57\% & 4.30\% \\
    \textsc{Qwen3.5-122B-A10B}
        & \textbf{30.40\%} & 0.27\% & \textbf{26.73\%}
        & 11.60\% & 0.96\% & 38.05\% \\
    \textsc{MiMo-V2.5-Pro}
        & \textbf{66.40\%} & 0.04\% & \textbf{14.43\%}
        & 43.40\% & 0.12\% & 23.40\% \\[2ex]

    & \multicolumn{3}{c}{\textbf{Concept}}
    & \multicolumn{3}{c}{\textbf{Keyword}} \\
    \cmidrule(lr){2-4}
    \cmidrule(lr){5-7}
    \textbf{GPQA}
    & \textbf{Msg.} & \textbf{Step} & \textbf{Bit}
    & \textbf{Msg.} & \textbf{Step} & \textbf{Bit} \\
    \midrule
    \textsc{gpt-oss-120b}
        & \textbf{98.99\%} & 0.00\% & \textbf{0.23\%}
        & 89.90\% & 0.00\% & 3.58\% \\
    \textsc{DeepSeek-V4-Flash}\footref{fn:deepseek-steps}
        & \textbf{89.39\%} & 2.53\% & \textbf{1.96\%}
        & 84.85\% & 3.22\% & 2.19\% \\
    \textsc{Qwen3.5-122B-A10B}
        & \textbf{91.92\%} & 0.12\% & \textbf{1.86\%}
        & 82.32\% & 0.35\% & 3.94\% \\
    \textsc{MiMo-V2.5-Pro}
        & \textbf{89.39\%} & 0.00\% & \textbf{2.44\%}
        & 76.26\% & 0.00\% & 7.32\% \\
    \end{tabular}
    \caption{
    End-to-end message accuracy, step-error, and bit-error rates after paraphrase for strategy \texttt{anticipate\_failure}. All quantities are percentages. Concept steganography consistently improves exact message recovery over the matched keyword baseline, and the gain is driven mainly by bit-error rather than step-error.
    }
    \label{tab:robustness}
\end{table*}

\subsection{Bit error dominates the robustness gap}
\label{sec:error-source}

A decode can fail in two ways: the paraphraser may merge or split steps (\emph{step error}), or it may scrub the conceptual/lexical signal within a step (\emph{bit error}). We do not instruct the paraphraser to preserve the step markers in the CoT, since in general we don't have control over a paraphrase defense and assume it will not be aware of a specific encoding strategy.
The Step and Bit columns in Table~\ref{tab:robustness} separate these effects. Across all rows, the concept channel attains a substantially \emph{lower} bit-error rate than the keyword channel, while step-error rates are small and similar. Robustness gains therefore mostly come from preserving \emph{within-step} signal under paraphrase, not from luckier segmentation. This pattern is consistent with the hypothesis that a content-preserving paraphraser preserves higher-level reasoning behavior while disrupting surface lexical cues.

\subsection{Errors are paraphrase-induced}
\label{sec:noise-floor}

The post-paraphrase error rates in Table~\ref{tab:robustness} could in principle be inflated by encoder hallucinations or decoder confusions that have nothing to do with the paraphrase itself. To rule this out, we run the decoder directly on the \emph{encoded} CoT, skipping the \textsc{Paraphrase} stage entirely (Appendix~\ref{sec:appendix-noise-floor}). Pre-paraphrase concept decoding stays at or above $95.5\%$ message accuracy with bit error at most $1.3\%$, and the keyword baseline is similarly tight. The bit-error increases reported in \S\ref{sec:error-source} are therefore attributable to the paraphraser rather than to encoder/decoder noise.

\subsection{Multiple channels can coexist}
\label{sec:multistrat}
We explore increasing channel capacity by combining multiple strategies in the same step. 
Concretely, Alice is asked to set strategy $s$ to $b^{(s)}_i$ for each strategy $s$ in the codebook, producing a single rewritten step that simultaneously realizes (or erases) each behavior. Decoders for different strategies then run independently against the paraphrased CoT. Table~\ref{tab:multistrat} (Appendix~\ref{sec:appendix-multistrat}) reports per-strategy decode quality when three strategies are encoded \emph{simultaneously} in the same CoT. The dominant strategy (\texttt{anticipate\_failure}) preserves much of its single-strategy capacity on the stronger models, while weaker strategies (\texttt{induction}, \texttt{constraint\_naming}) degrade substantially under parallel encoding ($30\text{--}46\%$ bit error in several cells). This indicates that capacity is not free across strategies: stacking bit-streams works best when carriers are semantically distinct, and falls off rapidly on strategies that are easier for the paraphraser to incidentally normalize.

\subsection{The channel transfers across models}
\label{sec:xmodel}
So far Alice (encoder), Eve (paraphraser), and Bob (decoder) are the same model. We ask whether the concept channel is idiosyncratic to that model by swapping one role to a held-out model, \textsc{MiMo-V2.5-Pro}, while reusing each source model's existing encoded/paraphrased CoTs (\texttt{anticipate\_failure}, seed 42). We consider the following two scenarios.

\begin{figure*}[t]
    \centering
    \includegraphics[width=\textwidth]{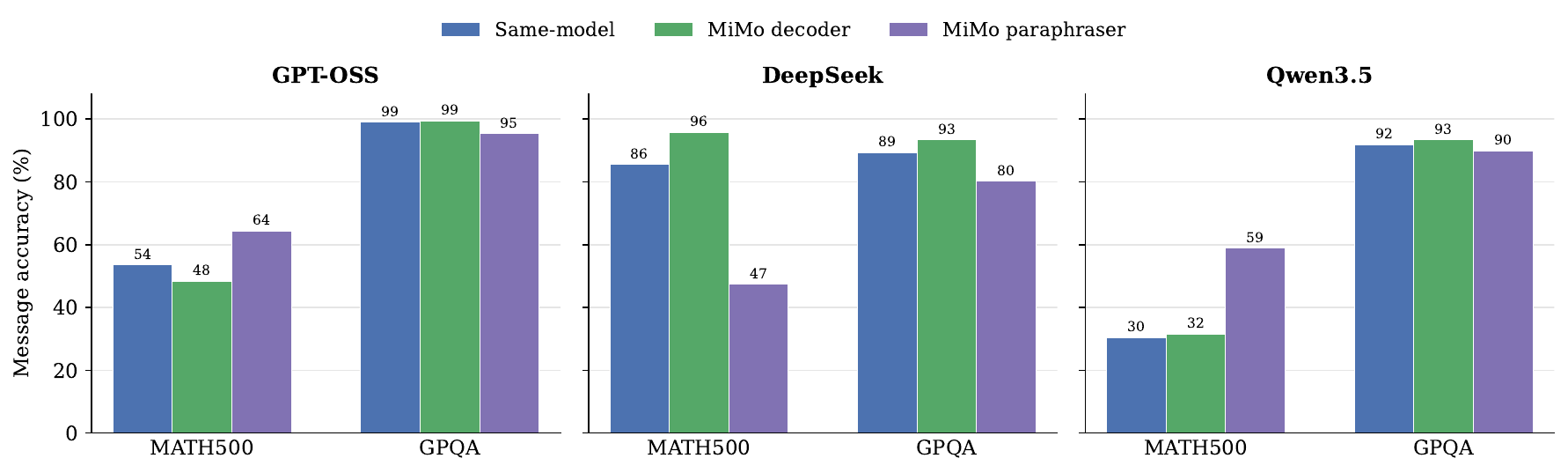}
    \caption{Cross-model transfer of the concept channel, by source model. Each source model runs the full encode pipeline under \texttt{anticipate\_failure} (seed 42). One role is then swapped to \textsc{MiMo-V2.5-Pro}. \textbf{Same-model}: original decoder and paraphraser. \textbf{MiMo decoder}: \textsc{MiMo} decodes the source model's paraphrased CoT. \textbf{MiMo paraphraser}: \textsc{MiMo} paraphrases the source model's encoded CoT, decoded by the source model. Bars are concept-channel message accuracy. \textsc{DeepSeek} uses the same step-count repair as Table~\ref{tab:robustness}, recomputed for all three conditions.}
    \label{fig:xmodel-transfer}
\end{figure*}

In the \emph{MiMo decoder} condition, \textsc{MiMo} is now confederate Bob, decoding the source model's paraphrased CoT (Eve unchanged); in the \emph{MiMo paraphraser} condition, \textsc{MiMo} is Eve, paraphrasing the source model's encoded CoT, which the source model then decodes. Figure~\ref{fig:xmodel-transfer} reports cross-model accuracy against the same-model baseline.

Swapping the \emph{decoder} barely moves recovery: a held-out decoder matches the original decoder to within a few points and often exceeds it (e.g.\ $+10.2$ for \textsc{DeepSeek} on MATH500), so the conceptual signal is legible, even to a model family that did not produce the encoding. Swapping the \emph{paraphraser} is more variable. On GPQA every cell stays within ${\sim}9$ points of the baseline, but on MATH500 the effect spans a large gain for \textsc{Qwen3.5} ($+28.6$) and \textsc{gpt-oss-120b} ($+10.8$) to a sharp drop for \textsc{DeepSeek} ($-38.2$). This asymmetry reinforces that the operative variable for robustness is the paraphraser, not the decoder model.

\footnotetext[\getrefnumber{fn:deepseek-steps}]{DeepSeek-V4-Flash's raw rewrite-stage step counts are affected by a likely tokenization-related counting issue that inflates the recorded step-error figures. The DeepSeek values in Table~\ref{tab:robustness} (and the multi-strategy DeepSeek values in Table~\ref{tab:multistrat}) incorporate a conservative prefix-only repair. The raw numbers, the failure mode, and a side-by-side comparison are reported in Appendix~\ref{sec:appendix-deepseek}.}

\begin{figure*}[t]
    \centering
    \includegraphics[width=\textwidth]{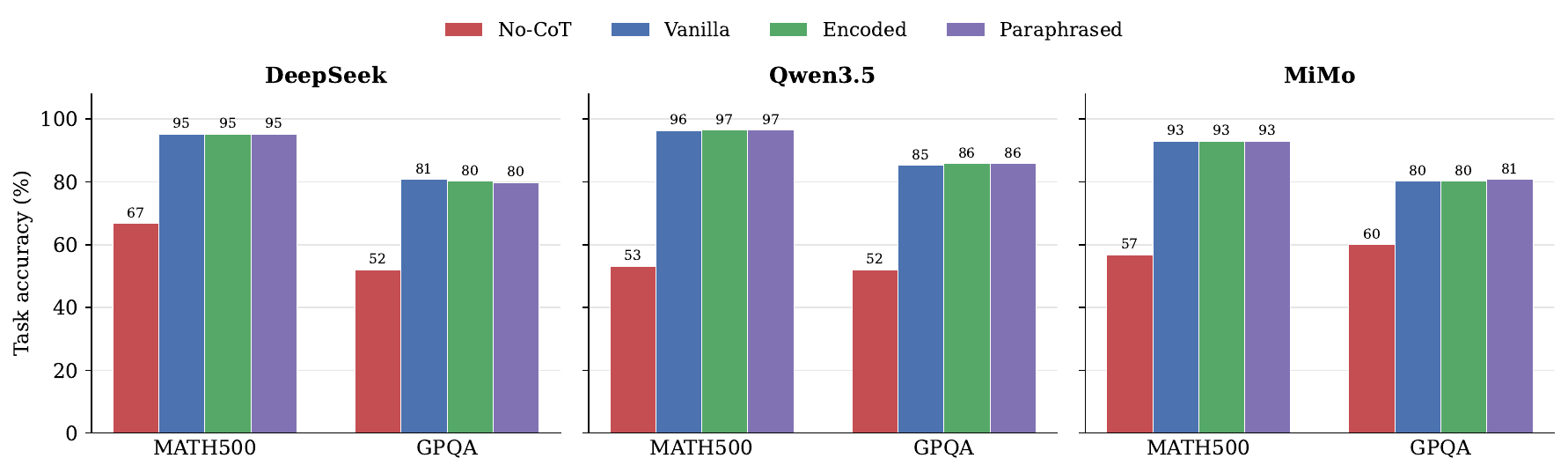}
    \caption{Task utility with and without CoT, by model. \textbf{No-CoT}: the model is prompted to emit only the final answer. \textbf{Vanilla / Encoded / Paraphrased}: an answer-only model receives the problem together with the source CoT after its explicit final-answer marker is stripped, and is instructed to derive the final answer from the partial reasoning. All Vanilla/Encoded/Paraphrased bars use \texttt{anticipate\_failure} with seed 42. Removing CoT collapses accuracy by 20--33 points on GPQA and 28--43 points on MATH500, while concept encoding and paraphrase keep accuracy within ${\sim}1\%$.}
    \label{tab:cot-utility}
\end{figure*}

\begin{table*}[t]
    \centering
    \small
    \begin{tabular}{llrrr}
    \textbf{Model} & \textbf{Paraphrase} & \textbf{Step Err.} & \textbf{Bit Err.} & \textbf{Msg. Acc.} \\
    \midrule
    \textsc{gpt-oss-120b}      & Standard      & 0.00\% & 0.23\%  & \textbf{0.9899} (196/198) \\
    \textsc{gpt-oss-120b}      & Strategy-aware & 0.00\% & 26.64\% & 0.4091 (81/198) \\[2ex]
    \textsc{Qwen3.5-122B-A10B} & Standard      & 0.12\% & 1.86\%  & \textbf{0.9192} (182/198) \\
    \textsc{Qwen3.5-122B-A10B} & Strategy-aware & 0.35\% & 35.03\% & 0.2020 (40/198) \\[2ex]
    \textsc{MiMo-V2.5-Pro}     & Standard      & 0.00\% & 2.44\%  & \textbf{0.8939} (177/198) \\
    \textsc{MiMo-V2.5-Pro}     & Strategy-aware & 0.42\% & 37.43\% & 0.1970 (39/198) \\
    \end{tabular}
    \caption{Strategy-aware paraphrasing as a targeted defense on GPQA with \texttt{anticipate\_failure}. Knowing the carrier strategy lets the paraphraser drive bit error up by an order of magnitude across all three tested models.}
    \label{tab:strat-aware}
\end{table*}

\subsection{Task utility is preserved}
\label{sec:utility}

A steganography channel is only interesting if it does not visibly degrade the underlying task.
Evaluation protocols extract the final answer from the CoT and compare it to the ground truth, so we can track how the answer changes across stages.
Table~\ref{tab:utility} (Appendix~\ref{sec:appendix-utility}) tracks task accuracy and per-stage answer-change rates from raw inference through paraphrase, using the full per-dataset record count as the denominator throughout (and treating records that lack an extractable final answer as wrong, and as a change relative to the prior stage). On cells where the upstream stage already commits to a final answer, encoding leaves it essentially untouched. Paraphrase introduces only modest accuracy drift, comparable between the concept and keyword variants, so the robustness gains in \S\ref{sec:robustness} do not come at the cost of utility.

\subsection{CoT utility is preserved}
\label{sec:cot-utility}

Two utility questions matter for a CoT-resident channel. \emph{First}, can the deployer afford to remove CoT altogether? The most direct defense against any CoT-resident channel is to forbid CoT itself: if there is no chain of thought, there is no surface to hide messages in. We test this by prompting each model to answer with only the final boxed answer (no reasoning), capping the response at 64 tokens on GPQA and 512 on MATH500. \emph{Second}, if CoT is kept but the pipeline rewrites it, does the rewritten CoT still carry enough reasoning for the final answer to be recoverable? For each source CoT, we strip the trailing answer line and ask the same model to output only the final answer, so the extractor must use the partial reasoning rather than copy the answer marker.

Figure~\ref{tab:cot-utility} reports both. Removing CoT collapses accuracy by 20--33 absolute points on GPQA and by 28--43 points on MATH500, so the deployer is structurally incentivized to keep CoT --- exactly the regime in which a paraphrase-robust CoT channel is a concern. Conditional on keeping CoT, encoding and paraphrase preserve answer recoverability from the rewritten reasoning: within each model/dataset block, the Vanilla, Encoded, and Paraphrased rows agree to within \(\sim 1\%\). Additionally, the fact that directly deriving the answer from the paraphrased CoT is almost as accurate as deriving it from the original CoT suggests that the paraphraser is \emph{sane} and preserves semantics.

\subsection{Strategy-aware paraphrasing as a defense}
\label{sec:mitigation}

Standard paraphrasing is content-preserving by design and therefore preserves the conceptual signal we encode. A targeted defense, however, can attempt to flatten specific reasoning behaviors --- e.g.\ instruct the paraphraser to rewrite each step so that no failure-anticipation phrasing remains. Table~\ref{tab:strat-aware} shows that this \emph{strategy-aware} paraphrase substantially degrades the concept channel on GPQA / \texttt{anticipate\_failure}: bit error rises by an order of magnitude and message accuracy drops by $58\text{--}72$ points. The channel is not eliminated, but a defender who knows (or can guess) which conceptual behavior is being used as the carrier has meaningful leverage. We view this as evidence that conceptual steganography is not unconditionally undetectable, and that defenses targeting reasoning content (rather than only surface form) are a constructive path forward.

\section{Conclusion}
\label{sec:conclusion}

We introduce \emph{conceptual steganography} --- a covert channel that hides bits in \emph{what reasoning behavior} a CoT step exhibits rather than in its words. It is substantially more robust to content-preserving paraphrasing than a matched keyword baseline across four open-weight reasoning models and two benchmarks. The encoded CoT preserves utility, and the final-answer accuracy of the surface form task remains largely intact throughout the pipeline. The defenses that we study reveals that scrubbing the surface form leaves the channel open, and only a strategy-aware paraphraser that targets the carrier behavior closes much of it, while removing CoT outright is too destructive on hard reasoning to be viable. Monitoring CoT for safety therefore has to read reasoning content, not just its phrasing. We hope this motivates further study of what behavior-level channels in CoT can be supervised.

\section*{Limitations}

\paragraph{Defender scope}
Our defender is a single LLM-based content-preserving paraphraser, plus a strategy-aware variant that is given the carrier behavior. We do not study training stronger defenders or detectors for our scenario.

\paragraph{Strategy-aware coverage}
The strategy-aware paraphrase results in Table~\ref{tab:strat-aware} cover one strategy (\texttt{anticipate\_failure}) on one dataset (GPQA) on three of the four models. We argue that the qualitative effect (codebook knowledge raises bit error by an order of magnitude) is the load-bearing observation, but we have not measured strategy-aware paraphrase on \texttt{induction}, \texttt{sanity\_check}, \texttt{constraint\_naming}, or on MATH-500.

\paragraph{No-CoT coverage}
Our no-CoT measurement (Figure~\ref{tab:cot-utility}) covers GPQA and MATH-500 on the three models for which the condition is well-defined, but the suppression is prompt-instructed rather than architectural: reasoning-tuned models may still emit some reasoning, and on MATH-500 a small fraction of DeepSeek and MiMo responses exceed the response cap mid-reasoning and are counted as wrong. The reported No-CoT accuracies on those cells are therefore lower bounds on prompt-suppressed-CoT capability.

\paragraph{LLM-judge decoding}
Decode accuracy is reported using an LLM judge, and bit-error and message-accuracy figures inherit any biases of that judge.

\paragraph{Stealth}
We measure decode accuracy (fraction of payload bits correctly recovered) and utility (final-answer accuracy) but not stealth: we do not build a discriminator that tries to detect whether a CoT is carrying a payload. Conceptual steganography may be more or less detectable than a keyword baseline along that axis, and we leave this open.

\paragraph{Capacity}
Our scheme transmits one bit per rewritten step using a single strategy at a time, or one bit per step per strategy when strategies are stacked. We do not characterize the information-theoretic capacity of behavior-level channels in CoT. Finer-grained or higher-rate encodings may exist.

\section*{Ethical considerations}

This work describes an attack: a way for an untrusted language model to hide a covert
payload in its chain of thought so that the message survives the paraphrase defense that
prior work treats as canonical. Publishing an attack always carries the risk of lowering
the bar for misuse. We judge the disclosure to be net-positive for three reasons. First,
the ingredients of the attack (segmenting a CoT, rewriting steps to exhibit or suppress a
reasoning behavior, decoding with a confederate model) are already available to any
capable actor, so the marginal capability we hand to an adversary is small. Second, the
defenders who rely on paraphrasing to sanitize CoT before human or automated inspection
are largely unaware that a content-preserving paraphraser is, by construction, blind to a
behavior-level carrier. Surfacing this gap is what lets defenders close it. Third, we do
not stop at the attack: we show that a strategy-aware paraphraser raises decoder bit error
by an order of magnitude, which points defenders at the axis that actually matters
(reasoning content, not surface form).

\paragraph{Use of datasets}
The two benchmarks we use are released under permissive licenses: GPQA~\citep{gpqa} is
distributed under the MIT License, and MATH-500, drawn from the MATH dataset~\citep{math500},
is likewise under the MIT License. Our use is non-commercial research use, consistent with the intended use of these resources. Both benchmarks contain no personally identifying or offensive content. Both datasets are in English.

\paragraph{Use of AI assistants}
The authors used AI-based coding and writing assistants while developing the experimental
pipeline and preparing the manuscript. Specifically, we used short-form input assistance for coding and writing. We also used AI to polish the language. For coding, we leveraged AI to scale up the experiment to more models and datasets starting from a human-written proof-of-concept pipeline. All scientific claims, experimental results, and
analyses were verified by the authors, who take full responsibility for the content.



\bibliography{custom}

\appendix

\section{Compute and Implementation}
\label{sec:appendix-compute}

All experiments are run through hosted APIs: we use the OpenRouter
API\footnote{\url{https://openrouter.ai/}} using the OpenAI Python client (\texttt{openai} 2.15.0). We therefore do not train or
host any model locally, and our own machine runs only lightweight orchestration and
metric computation on CPU. End-to-end wall-clock time for a
full (model, dataset) cell is typically under a day and is dominated by provider-side
queueing and rate limits rather than our own compute.

We use greedy decoding (temperature $0$) for every \textsc{Infer}, \textsc{Rewrite},
\textsc{Encode}, \textsc{Paraphrase}, and \textsc{Decode} call, and disable model
``thinking''/reasoning traces wherever the provider exposes that option, so that the CoT
we study is the visible response rather than a separate hidden reasoning stream.

All four models we study are open-weight and publicly available:
\textsc{gpt-oss-120b},\footnote{\url{https://huggingface.co/openai/gpt-oss-120b}}
\textsc{DeepSeek-V4-Flash},\footnote{\url{https://huggingface.co/deepseek-ai/DeepSeek-V4-Flash}}
\textsc{Qwen3.5-122B-A10B},\footnote{\url{https://huggingface.co/Qwen/Qwen3.5-122B-A10B}} and
\textsc{MiMo-V2.5-Pro}.\footnote{\url{https://huggingface.co/XiaomiMiMo/MiMo-V2.5-Pro}}

\section{DeepSeek-V4-Flash Step-Count Repair}
\label{sec:appendix-deepseek}

The DeepSeek-V4-Flash rows in Table~\ref{tab:robustness} and Table~\ref{tab:multistrat} report the \emph{repaired} measurements. Without repair, a small number of step-error figures are inflated by a likely tokenization-related counting issue in DeepSeek-V4-Flash's rewrite output rather than by paraphrase damage. This appendix isolates the issue, describes the conservative post-hoc repair applied to the main-text DeepSeek rows, and reports the raw numbers side-by-side with the repaired ones so that readers can verify both interpretations.

\subsection{Tokenization-related counting issue}
\label{sec:appendix-deepseek-parsing}

The rewrite stage asks the model how many step markers it produced and parses the answer with a regular-expression helper that returns the \emph{last} integer in the response:

\begin{quote}\small\ttfamily
matches = re.findall(r"\textbackslash d+", payload)\\
return int(matches[-1])
\end{quote}

This rule works reliably for the other models we evaluate. With DeepSeek-V4-Flash, however, the reported count is occasionally a spurious numeral (e.g.\ \texttt{55}, \texttt{44}, or \texttt{202}) even when only three to five visible step markers are present. When this happens, the recorded step count exceeds the true one and length-alignment metrics are inflated.

\subsection{Conservative prefix-only repair}
\label{sec:appendix-deepseek-repair}

We do not modify the runtime. The repair is a separate, post-hoc script that re-reads the saved artifacts. For each instance we (i) re-count the visible step markers in the rewritten CoT, (ii) take that as the ground-truth length when it is smaller than the recorded count, and (iii) truncate both the original and the decoded bit-strings to that length before recomputing step error, bit error, and message accuracy. Instances where the inflated count propagated forward (i.e.\ the encoder produced more steps than the visible markers indicate) are flagged and compared only on the intended prefix. No other model or stage is touched.

After repair, the most extreme step-error figures collapse: GPQA / concept drops from $23.35\%$ to $2.53\%$ and MATH500 / concept drops from $15.36\%$ to $1.40\%$, while bit-error and message-accuracy patterns are essentially unchanged.

\subsection{Raw vs.\ repaired numbers}
\label{sec:appendix-deepseek-tables}

Table~\ref{tab:appendix-deepseek-raw-vs-repaired} reports the raw DeepSeek-V4-Flash cells alongside the repaired counterparts that appear in Table~\ref{tab:robustness}; Table~\ref{tab:appendix-deepseek-multi} reports the multi-strategy variant under the same repair (the values reproduced in Table~\ref{tab:multistrat}).

\begin{table}[t]
\centering
\small
\setlength{\tabcolsep}{4pt}
\begin{tabular}{llrrr}
\textbf{Dataset} & \textbf{Channel} & \textbf{Step Err.} & \textbf{Bit Err.} & \textbf{Msg. Acc.} \\
\midrule
\multicolumn{5}{l}{\emph{Raw (pre-repair)}} \\
MATH500 & Concept & 15.36\% & 2.90\% & 0.7600 \\
MATH500 & Keyword & 15.68\% & 5.00\% & 0.6940 \\
GPQA    & Concept & 23.35\% & 3.10\% & 0.8232 \\
GPQA    & Keyword & 33.36\% & 1.59\% & 0.7727 \\[2ex]
\multicolumn{5}{l}{\emph{After step-count repair (as in Table~\ref{tab:robustness})}} \\
MATH500 & Concept & 1.40\%  & \textbf{2.53\%} & 0.8560 \\
MATH500 & Keyword & 0.57\%           & 4.30\% & 0.7780 \\
GPQA    & Concept & 2.53\%  & \textbf{1.96\%} & 0.8939 \\
GPQA    & Keyword & 3.22\%           & 2.19\% & 0.8485 \\
\end{tabular}
\caption{DeepSeek-V4-Flash single-strategy results before and after the prefix-only step-count repair. Bit-error and message-accuracy patterns are unchanged. The correction shows that almost all of the apparent step-error gap traces to DeepSeek-V4-Flash miscounting its own step markers, not to paraphrase damage.}
\label{tab:appendix-deepseek-raw-vs-repaired}
\end{table}

\begin{table}[t]
\centering
\small
\setlength{\tabcolsep}{4pt}
\begin{tabular}{lrrr}
\textbf{Strategy} & \textbf{Step Err.} & \textbf{Bit Err.} & \textbf{Msg. Acc.} \\
\midrule
\multicolumn{4}{l}{\emph{MATH500}} \\
\texttt{anticipate\_failure} & 0.68\% & \textbf{2.57\%}  & 0.8860 \\
\texttt{induction}           & 0.68\% & 36.63\%          & 0.1960 \\
\texttt{sanity\_check}       & 0.68\% & 8.83\%           & 0.6800 \\[2ex]
\multicolumn{4}{l}{\emph{GPQA}} \\
\texttt{anticipate\_failure} & 1.96\% & \textbf{1.04\%}  & 0.9293 \\
\texttt{constraint\_naming}  & 2.88\% & 31.42\%          & 0.2475 \\
\texttt{sanity\_check}       & 1.96\% & 20.25\%          & 0.4394 \\
\end{tabular}
\caption{DeepSeek-V4-Flash multi-strategy parallel decoding after the same repair. The strategy-specific bit-error pattern (dominant strategy preserves capacity, weaker strategies degrade) matches the cross-model trend in Table~\ref{tab:multistrat}.}
\label{tab:appendix-deepseek-multi}
\end{table}

\section{Pre-Paraphrase Noise Floor}
\label{sec:appendix-noise-floor}

To establish the noise floor referenced in \S\ref{sec:noise-floor}, we run the decoder directly on the \emph{encoded} CoT, skipping the \textsc{Paraphrase} stage entirely. Table~\ref{tab:noise-floor} reports this for the three open-weight models we control end-to-end. Pre-paraphrase concept decoding stays at or above $95.5\%$ message accuracy with bit error at most $1.3\%$, and the keyword baseline is similarly tight ($\geq 93.9\%$ msg.\ acc., $\leq 1.4\%$ bit error). The bit-error increases reported in \S\ref{sec:error-source} are therefore attributable to the paraphraser rather than to encoder/decoder noise.

\begin{table*}[t]
\centering
\small
\setlength{\tabcolsep}{6pt}
\begin{tabular}{lcccccc}
& \multicolumn{3}{c}{\textbf{Concept}}
& \multicolumn{3}{c}{\textbf{Keyword}} \\
\cmidrule(lr){2-4}\cmidrule(lr){5-7}
\textbf{MATH500}
& \textbf{Msg.} & \textbf{Step} & \textbf{Bit}
& \textbf{Msg.} & \textbf{Step} & \textbf{Bit} \\
\midrule
\textsc{gpt-oss-120b}      &  99.60\% & 0.04\% & 0.04\% & 99.20\% & 0.04\% & 0.13\% \\
\textsc{Qwen3.5-122B-A10B} &  99.00\% & 0.05\% & 0.18\% & 96.80\% & 0.18\% & 0.59\% \\
\textsc{MiMo-V2.5-Pro}     & 100.00\% & 0.00\% & 0.00\% & 95.80\% & 0.00\% & 0.85\% \\[2ex]
& \multicolumn{3}{c}{\textbf{Concept}}
& \multicolumn{3}{c}{\textbf{Keyword}} \\
\cmidrule(lr){2-4}\cmidrule(lr){5-7}
\textbf{GPQA}
& \textbf{Msg.} & \textbf{Step} & \textbf{Bit}
& \textbf{Msg.} & \textbf{Step} & \textbf{Bit} \\
\midrule
\textsc{gpt-oss-120b}      & 100.00\% & 0.00\% & 0.00\% & 95.45\% & 0.00\% & 1.38\% \\
\textsc{Qwen3.5-122B-A10B} &  95.45\% & 0.12\% & 1.28\% & 93.94\% & 0.35\% & 1.39\% \\
\textsc{MiMo-V2.5-Pro}     &  98.48\% & 0.00\% & 0.32\% & 94.95\% & 0.00\% & 1.06\% \\
\end{tabular}
\caption{Direct decoding results on the encoded CoT (no paraphrase). All quantities are percentages. Both concept and keyword channels recover the bit-string near-perfectly, so the post-paraphrase error in Table~\ref{tab:robustness} is attributable to the paraphraser, not to encoder/decoder noise. Strategy is \texttt{anticipate\_failure} throughout.}
\label{tab:noise-floor}
\end{table*}

\section{Multi-Strategy Parallel Decoding}
\label{sec:appendix-multistrat}

Table~\ref{tab:multistrat} accompanies the discussion in \S\ref{sec:multistrat} and reports per-strategy decode quality when three concept channels are encoded \emph{simultaneously} into the same CoT and decoded independently after paraphrase. The DeepSeek-V4-Flash rows incorporate the same prefix-only step-count repair described in Appendix~\ref{sec:appendix-deepseek} (raw counterparts in Table~\ref{tab:appendix-deepseek-multi}).

\begin{table}[t]
\centering
\small
\setlength{\tabcolsep}{5pt}
\begin{tabular}{llrr}
\textbf{Model} & \textbf{Strat.} & \textbf{Bit Err.} & \textbf{Msg. Acc.} \\
\midrule
\multicolumn{4}{l}{\emph{MATH500}} \\
\textsc{gpt-oss}  & AF & 22.70\% & 0.4800 \\
\textsc{gpt-oss}  & IN & 25.74\% & 0.3880 \\
\textsc{gpt-oss}  & SC & 16.03\% & 0.5340 \\[2ex]
\textsc{DeepSeek} & AF &  2.57\% & 0.8860 \\
\textsc{DeepSeek} & IN & 36.63\% & 0.1960 \\
\textsc{DeepSeek} & SC &  8.83\% & 0.6800 \\[2ex]
\textsc{Qwen3.5}  & AF & 37.18\% & 0.2000 \\
\textsc{Qwen3.5}  & IN & 45.76\% & 0.0860 \\
\textsc{Qwen3.5}  & SC & 29.06\% & 0.2800 \\[2ex]
\textsc{MiMo}     & AF & 22.83\% & 0.3860 \\
\textsc{MiMo}     & IN & 38.51\% & 0.2060 \\
\textsc{MiMo}     & SC & 16.41\% & 0.4660 \\[2ex]
\multicolumn{4}{l}{\emph{GPQA}} \\
\textsc{gpt-oss}  & AF &  0.23\% & 0.9899 \\
\textsc{gpt-oss}  & CN & 21.45\% & 0.3838 \\
\textsc{gpt-oss}  & SC &  8.07\% & 0.7222 \\[2ex]
\textsc{DeepSeek} & AF &  1.04\% & 0.9293 \\
\textsc{DeepSeek} & CN & 31.42\% & 0.2475 \\
\textsc{DeepSeek} & SC & 20.25\% & 0.4394 \\[2ex]
\textsc{Qwen3.5}  & AF &  7.19\% & 0.7222 \\
\textsc{Qwen3.5}  & CN & 36.08\% & 0.1717 \\
\textsc{Qwen3.5}  & SC & 19.26\% & 0.4646 \\[2ex]
\textsc{MiMo}     & AF & 12.09\% & 0.5960 \\
\textsc{MiMo}     & CN & 34.68\% & 0.1616 \\
\textsc{MiMo}     & SC & 28.42\% & 0.3030 \\
\end{tabular}
\caption{Multi-strategy parallel decoding: three concept channels are encoded into the same CoT and decoded independently after paraphrase. Strategies abbreviated: AF~=~\texttt{anticipate\_failure}, IN~=~\texttt{induction}, SC~=~\texttt{sanity\_check}, CN~=~\texttt{constraint\_naming}. Decode quality varies by strategy but stays well above chance on the dominant channel.}
\label{tab:multistrat}
\end{table}

\section{Task-Accuracy Preservation Across Pipeline Stages}
\label{sec:appendix-utility}

Table~\ref{tab:utility} accompanies the discussion in \S\ref{sec:utility} and reports task accuracy together with per-stage answer-change rates from raw inference through paraphrase, for every (model, dataset) cell. The complementary CoT-utility view (no-CoT vs.\ vanilla/encoded/paraphrased CoT as a reasoning source) is in Figure~\ref{tab:cot-utility} in the main text.

\begin{table*}[t]
\centering
\small
\begin{tabular}{lllrr}
\textbf{Model} & \textbf{Dataset} & \textbf{Stage / Variant} & \textbf{Task Acc.} & \textbf{$\Delta$} \\
\midrule
\textsc{gpt-oss}    & MATH500 & Infer                                  & 95.60\% & --- \\
\textsc{gpt-oss}    & MATH500 & Rewrite                                & 95.40\% & 0.20\% \\
\textsc{gpt-oss}    & MATH500 & Encode: concept \texttt{ant.\_failure} & 95.40\% & 0.00\% \\
\textsc{gpt-oss}    & MATH500 & Encode: keyword \texttt{ant.\_failure} & 95.40\% & 0.00\% \\
\textsc{gpt-oss}    & MATH500 & Paraphrase: concept                    & 92.60\% & 3.00\% \\
\textsc{gpt-oss}    & MATH500 & Paraphrase: keyword                    & 93.60\% & 2.40\% \\[2ex]
\textsc{DeepSeek}   & MATH500 & Infer                                  & 95.40\% & --- \\
\textsc{DeepSeek}   & MATH500 & Encode: concept                        & 95.00\% & 0.60\% \\
\textsc{DeepSeek}   & MATH500 & Paraphrase: concept                    & 95.00\% & 0.40\% \\
\textsc{DeepSeek}   & MATH500 & Paraphrase: keyword                    & 95.20\% & 0.00\% \\[2ex]
\textsc{Qwen3.5}    & MATH500 & Infer                                  & 96.40\% & --- \\
\textsc{Qwen3.5}    & MATH500 & Encode: concept                        & 96.40\% & 0.20\% \\
\textsc{Qwen3.5}    & MATH500 & Paraphrase: concept                    & 96.40\% & 0.00\% \\[2ex]
\textsc{MiMo}       & MATH500 & Infer                                  & 92.20\% & --- \\
\textsc{MiMo}       & MATH500 & Encode: concept                        & 93.00\% & 1.60\% \\
\textsc{MiMo}       & MATH500 & Paraphrase: concept                    & 93.00\% & 0.00\% \\
\textsc{MiMo}       & MATH500 & Paraphrase: keyword                    & 93.00\% & 0.00\% \\[2ex]
\textsc{gpt-oss}    & GPQA    & Infer                                  & 68.18\% & --- \\
\textsc{gpt-oss}    & GPQA    & Encode / Paraphrase (all variants)     & 68.18\% & 0.00\% \\
\textsc{DeepSeek}   & GPQA    & Infer                                  & 80.81\% & --- \\
\textsc{DeepSeek}   & GPQA    & Encode / Paraphrase (all variants)     & 80.30\% & 2.02\% \\
\textsc{Qwen3.5}    & GPQA    & Infer                                  & 84.34\% & --- \\
\textsc{Qwen3.5}    & GPQA    & Encode: concept                        & 85.35\% & 2.53\% \\
\textsc{Qwen3.5}    & GPQA    & Paraphrase: concept                    & 85.35\% & 0.00\% \\
\textsc{MiMo}       & GPQA    & Infer                                  & 79.80\% & --- \\
\textsc{MiMo}       & GPQA    & Encode / Paraphrase (all variants)     & 80.30\% & 1.52\% \\
\end{tabular}
\caption{Task-accuracy preservation across pipeline stages. $\Delta$ is the per-stage answer-change rate relative to the prior stage. Denominators use the full per-dataset record count throughout (500 for MATH500, 198 for GPQA). Records whose stage output lacks an extractable final answer count as wrong and as changed-vs-prior. Encoding induces little or no answer drift on cells where the prior stage already commits to a final answer.}
\label{tab:utility}
\end{table*}

\section{Single-strategy robustness across all four strategies}
\label{sec:appendix-single-strategy}

Table~\ref{tab:robustness} in the main text reports post-paraphrase robustness for a single carrier strategy (\texttt{anticipate\_failure}). Here we extend that measurement to all four strategies (\texttt{anticipate\_failure}, \texttt{induction}, \texttt{sanity\_check}, \texttt{constraint\_naming}) on \textsc{MiMo-V2.5-Pro}, under the identical pipeline: single-strategy concept encoding versus the matched keyword baseline, a standard content-preserving paraphrase, and concept-side decoding (seed 42). We report each strategy on the dataset where it naturally arises (Section~\ref{sec:methods}): \texttt{anticipate\_failure} and \texttt{sanity\_check} on both datasets, \texttt{induction} on MATH-500, and \texttt{constraint\_naming} on GPQA. The off-target pairings (\texttt{induction} on GPQA, \texttt{constraint\_naming} on MATH-500) are deferred to Appendix~\ref{sec:appendix-strategy-transfer}. The \texttt{anticipate\_failure} rows reproduce the corresponding cells of Table~\ref{tab:robustness}.

Table~\ref{tab:robustness-allstrat} shows that the concept channel beats the matched keyword baseline in every cell. The gain is again driven by lower bit-error rather than step-error, mirroring the main-text pattern. The absolute message-accuracy levels vary widely by strategy: \texttt{anticipate\_failure} is by far the strongest carrier, while \texttt{induction} and \texttt{sanity\_check} are weaker in absolute terms but they still improve over their keyword baselines.

\begin{table*}[t]
    \centering
    \small
    \setlength{\tabcolsep}{6pt}
    \begin{tabular}{lrrrrrr}
    & \multicolumn{3}{c}{\textbf{Concept}}
    & \multicolumn{3}{c}{\textbf{Keyword}} \\
    \cmidrule(lr){2-4}
    \cmidrule(lr){5-7}
    \textbf{GPQA}
    & \textbf{Msg.} & \textbf{Step} & \textbf{Bit}
    & \textbf{Msg.} & \textbf{Step} & \textbf{Bit} \\
    \midrule
    \texttt{anticipate\_failure}
        & \textbf{89.39\%} & 0.00\% & \textbf{2.44\%}
        & 76.26\% & 0.00\% & 7.32\% \\
    \texttt{sanity\_check}
        & \textbf{32.83\%} & 0.21\% & \textbf{24.92\%}
        & 20.71\% & 0.00\% & 30.54\% \\
    \texttt{constraint\_naming}
        & \textbf{16.16\%} & 2.76\% & \textbf{33.93\%}
        & 11.11\% & 2.33\% & 38.49\% \\[2ex]

    & \multicolumn{3}{c}{\textbf{Concept}}
    & \multicolumn{3}{c}{\textbf{Keyword}} \\
    \cmidrule(lr){2-4}
    \cmidrule(lr){5-7}
    \textbf{MATH500}
    & \textbf{Msg.} & \textbf{Step} & \textbf{Bit}
    & \textbf{Msg.} & \textbf{Step} & \textbf{Bit} \\
    \midrule
    \texttt{anticipate\_failure}
        & \textbf{66.40\%} & 0.04\% & \textbf{14.43\%}
        & 43.40\% & 0.12\% & 23.40\% \\
    \texttt{induction}
        & \textbf{31.80\%} & 1.47\% & \textbf{31.77\%}
        & 3.40\% & 0.35\% & 51.08\% \\
    \texttt{sanity\_check}
        & \textbf{37.80\%} & 0.23\% & \textbf{19.62\%}
        & 29.60\% & 0.70\% & 26.43\% \\
    \bottomrule
    \end{tabular}
    \caption{
    Single-strategy post-paraphrase robustness on \textsc{MiMo-V2.5-Pro} (seed 42), with each strategy reported on the dataset where it naturally arises (Section~\ref{sec:methods}). Columns mirror Table~\ref{tab:robustness}: end-to-end message accuracy, step-error, and bit-error after a standard content-preserving paraphrase, for the concept channel versus the matched keyword baseline. The \texttt{anticipate\_failure} rows reproduce the \textsc{MiMo} cells of Table~\ref{tab:robustness}. Bold marks the better channel per cell. Concept beats keyword in every cell.
    }
    \label{tab:robustness-allstrat}
\end{table*}

\section{Strategy transfer across domains}
\label{sec:appendix-strategy-transfer}

The strategies in Table~\ref{tab:robustness-allstrat} are each evaluated on the dataset where the underlying reasoning behavior naturally arises. A separate question is whether a strategy still functions as a carrier on a domain where its behavior is \emph{not} a natural fit: \texttt{induction} on GPQA-style scientific reasoning, and \texttt{constraint\_naming} on MATH-500 derivations. Table~\ref{tab:strategy-transfer} reports these two off-target pairings under the identical pipeline (\textsc{MiMo-V2.5-Pro}, seed 42).

The two off-target strategies behave very differently. \texttt{induction} transfers: even on GPQA, where inductive structure is comparatively rare, the concept channel still recovers far more messages than its keyword baseline ($31.3\%$ vs.\ $6.6\%$) at substantially lower bit error. \texttt{constraint\_naming} does not transfer to MATH-500: concept and keyword are statistically indistinguishable ($7.4\%$ vs.\ $8.2\%$ message accuracy, $42.5\%$ vs.\ $43.3\%$ bit error), both near the floor. When the carrier behavior is sparse in the domain, the paraphraser normalizes it just as easily in the concept channel as in the keyword channel, and the conceptual advantage disappears. This is consistent with the view that the concept channel's robustness comes from a reasoning behavior the paraphraser is inclined to preserve. Where that behavior is absent to begin with, there is nothing for the channel to ride on.

\begin{table*}[t]
    \centering
    \small
    \setlength{\tabcolsep}{6pt}
    \begin{tabular}{llrrrrrr}
    & & \multicolumn{3}{c}{\textbf{Concept}}
    & \multicolumn{3}{c}{\textbf{Keyword}} \\
    \cmidrule(lr){3-5}
    \cmidrule(lr){6-8}
    \textbf{Strategy} & \textbf{Dataset}
    & \textbf{Msg.} & \textbf{Step} & \textbf{Bit}
    & \textbf{Msg.} & \textbf{Step} & \textbf{Bit} \\
    \midrule
    \texttt{induction} & GPQA
        & \textbf{31.31\%} & 0.00\% & \textbf{33.09\%}
        & 6.57\% & 0.00\% & 46.98\% \\
    \texttt{constraint\_naming} & MATH500
        & 7.40\% & 2.24\% & \textbf{42.53\%}
        & \textbf{8.20\%} & 2.28\% & 43.27\% \\
    \bottomrule
    \end{tabular}
    \caption{
    Off-target strategy transfer on \textsc{MiMo-V2.5-Pro} (seed 42): each strategy evaluated on the dataset where its reasoning behavior is \emph{not} a natural fit. Columns match Table~\ref{tab:robustness-allstrat}. \texttt{induction} retains a large concept-over-keyword advantage even on GPQA, whereas \texttt{constraint\_naming} on MATH-500 collapses to parity with keyword, both near the floor.
    }
    \label{tab:strategy-transfer}
\end{table*}

\end{document}